\documentclass[article]{elsarticle}
\usepackage{hyperref}
\journal{Computer Methods and Programs in Biomedicine}
\usepackage{amsmath,amssymb,amsfonts}
\usepackage{algorithmic}
\usepackage{graphicx}
\usepackage{textcomp}
\usepackage{xcolor, colortbl}
\usepackage{geometry}
\newcommand{\norm}[1]{\left\lVert#1\right\rVert}
\newcommand{\intervalleoo}[2]{\mathopen{]}#1\,;#2\mathclose{[}}
\newcommand{\intervalleff}[2]{\mathopen{[}#1\,;#2\mathclose{]}}

\newcommand{\intervallefo}[2]{\mathopen{[}#1\,;#2\mathclose{[}}

\newcommand{\eP}{\mathcal{P}} % Ensemble P
\newcommand{\eM}{\mathcal{M}} % Ensemble M
\newcommand{\eA}{\mathcal{A}} % Ensemble A
\makeatletter
\newcommand{\tpmod}[1]{{\@displayfalse\pmod{#1}}}
\makeatother
\definecolor{Gray}{gray}{0.9}
\usepackage{xcolor, soul}
\sethlcolor{yellow}
\begin{document}
%%%%%%%%%%%%%%%%%%%%%%%%%%%%%%%%%%%%%%%%%%%%%%%%%%%%%%%%%%%%%%%%%%%%%%%%%
\begin{frontmatter}
%%%%%%%%%%%%%%%%%%%%%%%%%%%%%%%%%%%%%%%%%%%%%%%%%%%%%%%%%%%%%%%%%%%%%%%%%
\title{Three-dimensional reconstruction and characterization of bladder deformations}
% \address{Radarweg 29, Amsterdam}
% \fntext[myfootnote]{Since 1880.}

% \author[mymainaddress,mysecondaryaddress]{Elsevier Inc}

% \author[mysecondaryaddress]{Global Customer Service\corref{mycorrespondingauthor}}
% \cortext[mycorrespondingauthor]{Corresponding author}

% \address[mymainaddress]{1600 John F Kennedy Boulevard, Philadelphia}
% \address[mysecondaryaddress]{360 Park Avenue South, New York}
%%%%%%%%%%%%%%%%%%%%%%%%%%%%%%%%%%%%%%%%%%%%%%%%%%%%%%%%%%%%%%%%%%%%%%%%%
\author[address_lis]{Augustin C. Ogier\corref{mycorrespondingauthor}}
\cortext[mycorrespondingauthor]{Corresponding author}
\ead{augustin.ogier@gmail.com}

\author[address_crmbm]{Stanislas Rapacchi}
\ead{stanislas.rapacchi@univ-amu.fr}

\author[address_lis]{Marc-Emmanuel Bellemare}
\ead{marc-emmanuel.bellemare@univ-amu.fr}

\address[address_lis]{Aix Marseille Univ, Universite de Toulon, CNRS, LIS, Marseille, France}
\address[address_crmbm]{Aix Marseille Univ, CNRS, CRMBM, Marseille, France}
%%%%%%%%%%%%%%%%%%%%%%%%%%%%%%%%%%%%%%%%%%%%%%%%%%%%%%%%%%%%%%%%%%%%%%%%%

%%%%%%%%%%%%%%%%%%%%%%%%%%%%%%%%%%%%%%%%%%%%%%%%%%%%%%%%%%%%%%%%%%%%%%%%%
\begin{abstract}
\textbf{Background and Objective:} Pelvic floor disorders are prevalent diseases and patient care remains difficult as the dynamics of the pelvic floor remains poorly known. So far, only 2D dynamic observations of straining exercises at excretion are available in the clinics and the understanding of three-dimensional pelvic organs mechanical defects is not yet achievable. In this context, we proposed a complete methodology for the 3D representation of the non-reversible bladder deformations during exercises, directly combined with synthesized 3D representation of the location of the highest strain areas on the organ surface.
\textbf{Methods:} Novel image segmentation and registration approaches have been combined with three geometrical configurations of up-to-date rapid dynamic multi-slices MRI acquisition for the reconstruction of real-time dynamic bladder volumes.
\textbf{Results:} For the first time, we proposed real-time 3D deformation fields of the bladder under strain from in-bore forced breathing exercises. The potential of our method was assessed on eight control subjects undergoing forced breathing exercises. We obtained average volume deviation of the reconstructed dynamic volume of bladders around 2.5\% and high registration accuracy with mean distance values of  0.4 $\pm$ 0.3 mm and Hausdorff distance values of 2.2 $\pm$ 1.1 mm. 
\textbf{Conclusions:} Immediately transferable to the clinics with rapid acquisitions, the proposed framework represents a real advance in the field of pelvic floor disorders as it provides, for the first time, a proper 3D+t spatial tracking of bladder non-reversible deformations. This work is intended to be extended to patients with cavities filling and excretion to better characterize the degree of severity of pelvic floor pathologies for diagnostic assistance or in preoperative surgical planning.
\end{abstract}
%%%%%%%%%%%%%%%%%%%%%%%%%%%%%%%%%%%%%%%%%%%%%%%%%%%%%%%%%%%%%%%%%%%%%%%%%
%%%%%%%%%%%%%%%%%%%%%%%%%%%%%%%%%%%%%%%%%%%%%%%%%%%%%%%%%%%%%%%%%%%%%%%%%
\end{frontmatter}
%%%%%%%%%%%%%%%%%%%%%%%%%%%%%%%%%%%%%%%%%%%%%%%%%%%%%%%%%%%%%%%%%%%%%%%%%

%%%%%%%%%%%%%%%%%%%%%%%%%%%%%%%%%%%%%%%%%%%%%%%%%%%%%%%%%%%%%%%%%%%%%%%%%
\section{Introduction}

Pelvic floor disorders affect one fourth of adult women and the prevalence increases with age, up to 50\% for women older than 80 years \cite{nygaard2008prevalence,wu2014prevalence}. Some experts refer to them as hidden epidemics to describe the extent of these diseases and their importance in the population \cite{delancey2005hidden}. These pathologies result from various factors such as aging, pregnancy, childbirth, obesity, injuries \cite{gurland2021consensus} and are mainly characterized by a weakening of the pelvic floor muscles and ligaments leading to mechanical dysfunction of the pelvic organs support structures. Symptoms range from urinary and bowel incontinence, sexual dysfunction to abnormal descent of pelvic organs.

Besides the clinical examination, magnetic resonance imaging (MRI) has emerged as one of the most attractive non-invasive methods for pelvic diagnosis \cite{woodfield2010imaging}. Dynamic MRI examinations of a pelvic strain exercise are considered essential \cite{el2017magnetic} for the investigation of the pelvic area and guidelines emphasize their importance to identify pathological organs deformations. 
Current clinical practice involves 2D dynamic MRI acquiring a single sagittal plane during straining exercises involving excretion to qualitatively assess the displacement and deformations of the main pelvic organs (bladder, uterus-vagina, rectum). During visual inspection, radiologists assess the displacement of the pelvic organs relative to bone-anchored lines, such as the pubococcygeal line \cite{zhang2022dynamic}. Beyond manual measurements, automated measures of strain-induced deformation features were proposed to build a quantitative characterization of the pelvic organ dynamics to distinguish pathological cases from healthy ones \cite{rahim2013diffeomorphic} and to graduate the severity of pelvic organ prolapses \cite{nekooeimehr2018automated}.

The visualization and characterization of pelvic dynamics by MRI has so far only been studied in the midsagittal plane. Although the majority of pelvic movements are identifiable in this plane, an understanding of three-dimensional deformities could provide full assessment of the pelvic organs mechanical defects. Due to speed limitation in MRI acquisition, 3D imaging have been exclusive to static pelvic MRI so far, to observe anatomy at rest during instructed apnea. 
Ultrasound have been used to observe organs during loading exercises \cite{salsi2017three}, such as Valsalva maneuver for the diagnosis of prolapse, but remains limited due to image interpretation difficulties and restricted field of view. Although significant research has been performed, the pelviperineal physiology and the anatomic basis of pelvic floor diseases remain unclear \cite{weber2005pelvic, maher2013surgical}.

This misunderstanding is the main cause for repair surgery relapse \cite{abed2011incidence}. To better comprehend the complete mechanical functioning of the pelvic floor and to provide better post-operative support, biomechanical models have been proposed in recent years with the aim of providing patient-specific 3D simulations of the pelvic region \cite{bellemare2007toward, chen2015female}. Although pelvic system geometry could be known from high-resolution MRI images, patient-specific boundary conditions and material biomechanical properties are impossible to probe \emph{in vivo} as it requires invasive procedures \cite{rubod2012biomechanical}. Biomechanical numerical simulation could offer indirect characterization of pelvic organs mechanical impairment and guide surgical intervention, anticipating patients recovery.
Based on shape descriptors and anatomical references approximation, comparison criteria between simulation sequence and 2D midsagittal dynamic MRI of a same patient was proposed to validate and enhance a physical model \cite{rahim2011quantiative}. 2D dynamic data are not sufficient to depict realistic 3D biomechanical models. Guiding 3D biomechanical models through a single 2D slice fails to fully simulate pelvic organs dynamics as simulation errors remain in off-plane areas \cite{courtecuisse2020three}. Recent attempts to provide a finite elements based simulation of organ behaviours still lack realistic patient 3D data \cite{gordon2019framework}.

In light of the current state-of-the art in the study of pelvic disorders, there is a definite need for methods to generate dynamic 3D volumes of pelvic organs. Such approach would allow full visualization of the 3D dynamics of the organs in order to help diagnosis and open the way to the geometric characterization of real 3D deformations of the organs. A complete 3D characterization of the pelvic floor would provide a better classification of pathologies but also feed biomechanical simulators that could improve the surgical management of pelvic organ prolapse.

Pelvis straining MRI falls within real-time applications along with bolus-tracking perfusion MRI since organs do not recover the same position after each strain exercises and intra-cavities fluids can be excreted \cite{el2017magnetic}. Patients cannot voluntarily strain their pelvis with the same intensity and regularity during repetitive cycles of forced breathing. Current state-of-the-art perfusion MRI methods rely either on multi-slices (3 to 6) dynamic acquisitions in cardiac perfusion MRI, with an update rate of one heartbeat ($\approx$ 1 s) \cite{mcelroy2020combined} or 3D dynamic acquisitions such as liver perfusion MRI, with an update rate of few seconds per volume at best \cite{feng2016xd}. The targeted spatial resolution combined with the extreme nature of the pelvic organ unloading induced during these exercises requires an order of magnitude faster imaging ($<$ 1 s) and prevents 3D real-time pelvis MRI.

As mid-sagittal acquisitions, at about 1 frame per second, is the consensus for exploration of the lesser pelvis region\cite{el2017magnetic}, we have kept these features in our imaging schemes. In particular, we focused on maintaining the frame rate of one image per second. However, to gain 3D information, it was necessary to complete the spatial information, but with a short temporal footprint to minimize errors due to deformations. For this purpose, we tested 3 spatial multi-slices configurations among the most obvious ones, that can be easily reproduced in clinical set-ups. We compared them to estimate the errors on organs tracking induced by privileging spatial coverage or spatial resolution. This shall give us an idea of the best configuration to use to see the bladder but also possibly the main neighboring organs. Paired with a 3D static acquisition, the proposed acquisition aims to provide volumetric dynamic reconstruction.
This study was a complete extension of a previous seminal work \cite{ogier20193d}.

Initially the bladder was the main organ concerned as it is straightforward and the simplest organ to consider being homologous to the sphere. However we have described our methodology in a generic way as we intend to apply it in future studies for the visualization of the uterus and rectum.
We leveraged a semi-automated segmentation method, based on a fusion of image registration approaches, for the tracking of organs in the dynamic planes, regardless of their spatial orientation. 
Linear interpolations on the geodesic path of each multi-planar segmentation have been proposed to refine the time scale inherited from the acquisition methods. 
Finally, we built dynamic 3D representations of organs through non-linear geometric registrations between each dynamic partial volumes and the complete static volume of the high-resolution static acquisition. 
As this study produced dynamic 3D bladder volumes that can't be directly compared to a ground truth impossible to acquire, we emphasized metrics to assess the adequacy of the volume reconstructions: accuracy of the registration and conservation of the bladder volume.

To our knowledge, this study is the first to propose both dynamic 3D visualization of the pelvic region during strain exercise and a dynamic 3D tracking of the organs. Furthermore, the organ reconstruction process allowed a synthesized characterization of the 3D deformations undergone by the pelvic organs during loading exercises. Analysis of the deformation fields allowing the reconstruction of organs, and more particularly the study of the resulting Jacobians, revealed local volume changes over time and provided a high-level 3D representation of the location of the highest strain areas on the organs surface. Specifically, analysis of Jacobians validated that our complete reconstruction methodology produced similar results between different geometric configurations of dynamic acquisitions.

%%%%%%%%%%%%%%%%%%%%%%%%%%%%%%%%%%%%%%%%%%%%%%%%%%%%%%%%%%%%%%%%%%%%%%%%%
% Methods
%%%%%%%%%%%%%%%%%%%%%%%%%%%%%%%%%%%%%%%%%%%%%%%%%%%%%%%%%%%%%%%%%%%%%%%%%
\section{Methods}

%%%%%%%%%%%%%%%%%%%%%%%%%%%%%%%%%%%%%%%%%%%%%%%%%%%%%%%%%%%%%%%%%%%%%%%%%
\subsection{MRI acquisitions}

Both static and dynamic acquisition methods implemented in the context of this study were performed using a 1.5T MRI scanner (MAGNETOM Avanto, Siemens AG, Healthcare Sector, Erlangen, Germany) using a spine/phased array coil combination and $T_{1}/T_{2}$ weighted balanced steady-state free precession sequences ($T_{1}/T_{2}W$ bSSFP). In comparison to the $T_{2}W$ sequence commonly used for pelvic area MRI, the $T_{1}/T_{2}W$ bSSFP sequence allows an optimal acquisition speed combined with a strong contrast between body fluids and the different pelvic organs tissues: uterus in dark gray, bladder in white (hyper-signal), and rectum and viscera (dark gray tissue and light gray content). 

\begin{figure}[!b]
\centerline{\includegraphics[width=0.7\columnwidth]{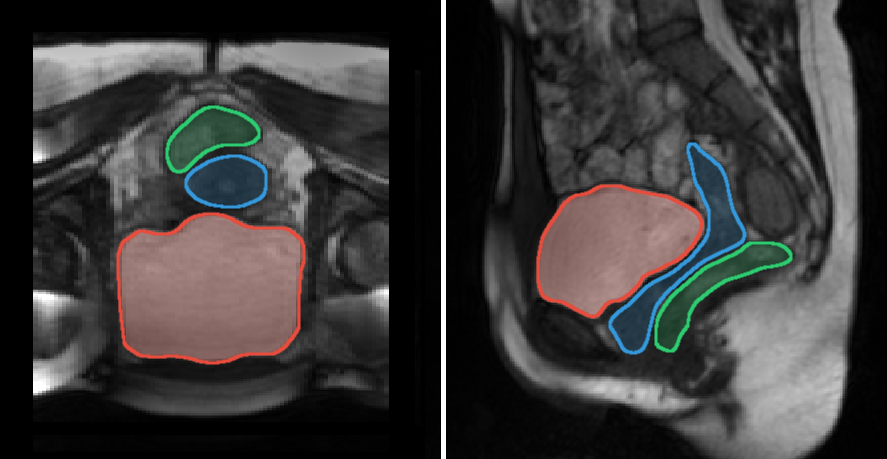}}
\caption{Axial (left) and sagittal (right) views of pelvic floor acquired during a maximum expiration apnea with delineations of the major pelvic organs: bladder (red), uterus/vagina (blue), and rectum (green).}
\label{CM_static_ebauche}
\end{figure}

\subsubsection{Static acquisition configuration}
A quasi-isotropic 3D static image was acquired, as depicted in Fig.~\ref{CM_static_ebauche}. High resolution in the sagittal plane and high contrast allowed the anatomy of the pelvic region to be precisely delineated. The 3D T$_{1}$/T$_{2}$W bSSFP image (TR = 2.21 ms, echo time: 0.99 ms, flip angle: 32$^{\circ}$, field of view: 277 mm $\times$ 320 mm $\times$ 88 mm, voxel size: 0.83 mm $\times$ 0.83 mm $\times$ 2.0 mm, GRAPPA~=~2) was recorded during a maximum expiration apnea of 18 seconds. Static acquisition was performed to allow precise three-dimensional manual segmentation of the pelvic organs used as references for the reconstruction scheme detailed in section \ref{section_3drec}.

\subsubsection{Dynamic acquisition configurations}
Dynamic sequences provided a lower spatial resolution and a reduced contrast compared to static sequences, but allowed the monitoring of pelvic dynamics during organ loading exercises. Analysis of the biomechanical deformations induced during these exercises should allow to identify elasticity defects in pelvic floor tissues.
Three-dimensional dynamic MRI can be acquired either using slab excitations or multi-planar slice excitations. Due to the extended temporal footprint of 3D excitations ($>$1s), blurring can occur when observing rapid deformations. Thus, fast acquisition ($\approx$ 100 ms) of multi-planar 2D slices was preferred to follow the motion of the pelvic organs. In this study, we sought to harness the potential of high-density coil arrays to accelerate MRI using parallel imaging techniques.

\begin{figure}[!t]
\centerline{\includegraphics[width=0.7\columnwidth]{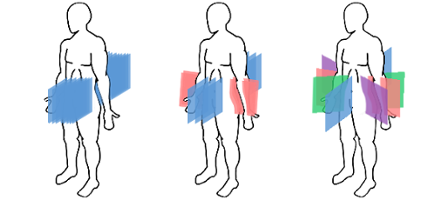}}
\vspace{0.5em}
\begin{minipage}[b]{.33\linewidth} \centering \centerline{$C_{lines}$} \medskip \end{minipage}
\begin{minipage}[b]{.32\linewidth} \centering \centerline{$C_{grid}$} \medskip \end{minipage}
\begin{minipage}[b]{.33\linewidth} \centering \centerline{$C_{star}$} \medskip \end{minipage}
\vspace{-1.8em}
\caption{Geometrical configuration of the planes for the three dynamic multi-planar acquisitions sequences.}
\vspace{-0.5em}
\label{stickPlanes}
\end{figure}

Non-linear reconstruction techniques, such as Compressed Sensing \cite{lustig2007sparse}, were not considered in this preliminary work to maintain reliability of the imaging information and clinical transferability. The sliding-window temporal-GRAPPA (TGRAPPA) technique \cite{breuer2005dynamic} allowed for a significant acceleration. In TGRAPPA, consecutive images share low-frequency information to provide basis for image reconstruction. Assuming the pelvic deformations do not modify the low-frequency image information within the range of a second, the full coverage of the pelvic region was guaranteed using up to a maximum of 12 slices, regardless of the geometry configurations, within a second. Another interesting feature of this multi-planar setup is the T$_1$-recovery of the signal between images, which boosts signal-to-noise ratio and offers a reinforced contrast between tissues and liquids.

\subsubsection{Spatial order of dynamic acquisitions}
Three dynamic sequences in multi-planar configurations were acquired, with three different geometric configurations that varied plane number and locations but maintained the time for acquiring one set of all planes within 1s. We denoted the different geometries as $C_{star}$, $C_{grid}$ and $C_{lines}$ (cf. Fig.~\ref{stickPlanes}). %The $Star$ configuration consisted of one sagittal plane, one coronal plane and two others planes oriented at 45 degrees in the axial plane in regard to the two others. 
The $Star$ configuration consisted of one coronal plane, the mid-sagittal plane and 2 others oriented at about 45 degrees to the sagittal. The $Grid$ configuration featured three sagittal and two coronal planes. The $Lines$ configuration was made of several parallel sagittal planes covering the body from the right lateral side to the left lateral side. Whatever the configuration, the number of planes of one acquisition cycle was noted $N_p$. For $C_{star}$ and $C_{grid}$, $N_p$ was equal to 4 and 5, respectively. The number of planes for $C_{lines}$ was defined according to the corpulence of each subject to cover the pelvic area as much as possible and was typically set at 10 ($\pm$ 2).

\begin{table}[!htbp]
\centering
\caption{Acquisition parameters of the three dynamic MRI sequences}
%\small
\setlength{\tabcolsep}{3pt}
\begin{tabular}{|p{120pt}|p{60pt}|p{60pt}|p{60pt}|}
\hline
\rowcolor{Gray}
Configuration               & $C_{star}$    & $C_{grid}$    & $C_{lines}$   \\ \hline
Slice thickness (mm)        & 5             & 6             & 4             \\
\rowcolor{Gray}
Number of cycles            & 100           & 100           & 60            \\
Time repetition (ms)        & 2.9           & 2.6           & 1.9           \\
\rowcolor{Gray}
Slice acquistion time (ms)  & 183.72        & 124.95        & 91.61        \\
Field of view (mm)          & 302 $\times$ 350     & 300 $\times$ 350     & 299 $\times$ 350     \\
\rowcolor{Gray}
In-plane resolution (mm)    & 1.09 $\times$ 1.09   & 1.36 $\times$ 1.36   & 1.82 $\times$ 1.82   \\ \hline
\end{tabular}
\label{paramMRIdyn}
\end{table}

\subsubsection{Temporal order of dynamic acquisitions}
\begin{figure}[!t]
\centerline{\includegraphics[width=0.8\columnwidth]{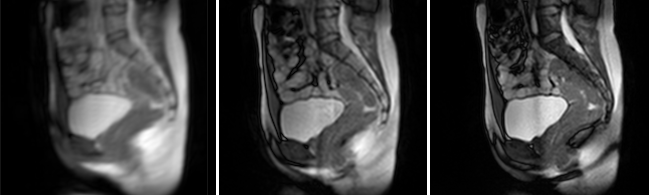}}
\vspace{0.5em}
\begin{minipage}[b]{.33\linewidth} \centering \centerline{$C_{lines}$} \medskip \end{minipage}
\begin{minipage}[b]{.32\linewidth} \centering \centerline{$C_{grid}$} \medskip \end{minipage}
\begin{minipage}[b]{.33\linewidth} \centering \centerline{$C_{star}$} \medskip \end{minipage}
\vspace{-1.8em}
\caption{Difference in contrast and resolution between similarly located sagittal planes from the different dynamic acquisition configurations.}
\label{sagittalViewEx}
\end{figure}

Magnetic resonance imaging involves a close relationship between spatial resolution, temporal resolution and contrast of the acquired image. This leads to a trade-off where modification of one of these parameters induces a change in the others. Increasing the duration of a single image plane acquisition (\emph{i.e.} the sequence-specific slice acquisition time [TS]) resulted in an acquired image with improved contrast and better spatial resolution. To follow the motions of the pelvic organs, an acquisition cycle, which acquired all the planes one after the other, had to be acquired in less than a second. Thus, defining the timeframe of a cycle by $ N_p \cdot \mathrm{TS}$, each plane of a given configuration had to be acquired in less than $1/N_p$ seconds. 
Therefore, assuming the same field of view, the lower the $N_p$ required by the geometric configuration, the longer the $\mathrm{TS}$ was, and therefore the more precise the spatial resolution, as detailed in the table \ref{paramMRIdyn}, and better the contrast, as illustrated in Fig.~\ref{sagittalViewEx}. To allow for clinical-grade signal-to-noise ratio, the $\mathrm{TS}$ of $C_{Lines}$ was set at a minimum of 110ms, even with a $N_p > 10$.

The number of cycles, $N_c$, acquired for each dynamic configuration was determined to set the total acquisition time (\emph{i.e.} $N_p \cdot \mathrm{TS} \cdot N_c$ ) around 1:20 min in order to record several deformation phases of pelvic organs during loading exercises.

Regardless of dynamic configuration, for each $p \in \intervallefo{0}{N_p}$, the spatial indexes of acquisition planes, we denoted $\eP_p$ the set of the acquired planes $\{\eP^k_p\}$ with $k \in \intervallefo{0}{N_c}$ the temporal index of the acquisition cycle.

%%%%%%%%%%%%%%%%%%%%%%%%%%%%%%%%%%%%%%%%%%%%%%%%%%%%%%%%%%%%%%%%%%%%%%%%%
\subsection{Mathematical formalism}

We introduce here the mathematical formalism of diffeomorphic registration widely used in the following sections. A diffeomorphism, $\phi$, allows to transform a target image \emph{I} towards a source image \emph{J}, and to find the correspondence between these two shapes, such as $\phi_x^t I \approx J$. The shape I is warped into the new coordinate system by $\phi_x^t I = I \circ \phi(x,t)$, where the geodesic path $\phi(x,t)$ is described in \cite{beg2005computing}. 
The diffeomorphic function, $\phi(x,t)$, parameterized by time, $t \in [0,1]$ and $x$, spatial coordinate, is defined by $\delta \phi^t_x / \delta t = v_t(\phi^t_x)$ and $\phi^0_x$ = \textbf{Id}, with $v(x,t)$ a time-dependent velocity field.

%%%%%%%%%%%%%%%%%%%%%%%%%%%%%%%%%%%%%%%%%%%%%%%%%%%%%%%%%%%%%%%%%%%%%%%%%
\subsection{Pelvic organ segmentation}

\subsubsection{State of the art}
Automatic segmentation of main pelvic organs is a challenging task due to overlapping MRI contrasts between the uterus, vagina, rectum and surrounding soft tissues. Only the bladder is natively contrasted due to the liquid it contains. To enhance the contrast of the other organs, extrageneous liquid could be inserted in the pelvic cavities, but the clinical procedure loses its non-invasive nature. 

Several methods have been proposed for the segmentation of pelvic organs using MRI but mainly on 3D static axial images \cite{pasquier2007automatic,duan2010coupled,ma2013segmentation} and only a few semi-automatic methods have already been developed for segmentation of dynamic 2D sagittal T$_{2}$W MRI images \cite{namias2014uterus,jiang2015b,nekooeimehr2018automated}. Although promising, these methods still require manual initialization and have not been able to obtain segmentations with a average Dice similarity coefficient (DSC) greater than 0.90 for the bladder \cite{namias2014uterus,nekooeimehr2018automated} or with a mean Hausdorff distance lower than 9 mm \cite{jiang2015b}. 
Recently, convolutional networks have shown good results in addressing the complex segmentation of pelvic organs in MRI but mostly with static volumes\cite{nie2018strainet}. 
Overall, the segmentation of multi 2D dynamic MRI datasets is still challenging due to the combined non-rigid deformations and natively depleted contrasts.
As a result, the segmentation process of pelvic organs has mainly been performed manually or semi-automatically in studies that have focused on pelvic dynamics \cite{rahim2013diffeomorphic, courtecuisse2020three}. 

\subsubsection{Our method}
In this study, the segmentation stage is the first step of a complete process allowing characterization of pelvic organ deformations (cf. Fig.~\ref{pipeline}). This step is crucial, as incorrect segmentations would lead to false deformation estimations. So, for the segmentation of pelvic organs, a semi-automated segmentation method was proposed on each set of 2D dynamic planes $\eP_p$ independently, to balance between high segmentation accuracy and user's dedication. For each set of acquired planes $\eP_p$, segmentation masks $\eM_p$ were manually depicted on a few slices only, at regular temporal intervals. The contours were then automatically propagated over all the remaining frames according to the method described in \cite{ogier2017individual}. The manually segmented slices were denoted ${\eM_p}^{i \cdot n}$, with $i \in \mathbb{N}$, and $n$ the time step of the mandatory manual segmentation. According to our method, the manual segmentation of the slices ${\eM_p}^{i \cdot n}$ and ${\eM_p}^{(i + 1) \cdot n}$ allowed the automatic segmentation of the $n - 1$ slices of the interval using a fusion of different propagation schemes.

The propagation from one manually segmented slice ${\eM_p}^{i \cdot n}$ to the next ${\eM_p}^{(i + 1) \cdot n}$ could be achieved, from an Eulerian perspective, by successively estimating the correspondence maps between the $n - 1$ grey level $\eP_p$ slices, pairwise. This process could be done in both forward and backward strategies depending on the starting slices. The forward strategy sought to estimate the diffeomorphism $\phi^k_{E_f}$, given by: \begin{equation} \phi^{k}_{E_f} = \phi^{i \cdot n + 1}_{v_f} \circ \phi^{i \cdot n + 2}_{v_f} \circ \dotsc \circ \phi^{k}_{v_f} \end{equation} where $\phi^{j}_{v_f}$ is the deformation field mapping $\eP^{j-1}_p$ to $\eP^{j}_p$. In a similar way, the backward strategy was to estimate $\phi^k_{E_b}$, given by the combination of the diffeomorphism $\phi^{j}_{v_b}$, mapping $\eP^{j+1}_p$ to $\eP^j_p$. Therefore, for $k \in \intervalleoo{i \cdot n}{(i+1) \cdot n}$ with $i \in \mathbb{N}$, the propagation to the missing segmentations $M^k_p$ could have been performed by applying either $\phi^{k}_{E_f} \eM^{i \cdot n}_p$ or $\phi^{k}_{E_b} \eM^{(i+1) \cdot n}_p$. 
As registrations were performed between successive MRI slices, these Eulerian approaches did follow anatomical variations but rapidly diverged given that errors were accumulating with the number of $\phi^{j}_v$ combinations. So, the accuracy decreased with respect to the distance from the starting slice. To minimize this divergence effect, the two approaches were merged and the missing segmentations were generated by: \begin{equation} \eM^k_p = \left(\alpha_k \phi^k_{E_f} + \beta_k (\phi^k_{E_b} \circ \phi_{L_f}) \right) M^{i \cdot n}_p \end{equation} with $\phi_{L_f}$, the deformation field resulting from the geometric registration between the two initial manually segmented masks. 
This diffeomorphism $\phi_{L_f}$ allowed to perform a reference frame shift, and merge the two Eulerian approaches. The determination of coefficients $\alpha_k$ and $\beta_k$ were ruled by an $\arctan$ function to keep the high accuracy of each Eulerian propagation process near its propagation starting slice. The complete algorithm of the merging process and parametric details of the diffeomorphic registrations are available in \cite{ogier2017individual}. All the segmentations were visually checked at the end of the process to ensure their quality.

%%%%%%%%%%%%%%%%%%%%%%%%%%%%%%%%%%%%%%%%%%%%%%%%%%%%%%%%%%%%%%%%%%%%%%%%%
\subsection{Temporal reconstruction of segmentations} \label{section_tempRec}

After the segmentation process, $N_p$ datasets $\eM_p^k$ were obtained, each with a cardinality of $N_c$. The next step towards the 3D reconstruction was to merge all $M_p$ datasets into a unique set, $\eA$, that we represented as a 2D matrix $[(t,p)]$. Taking $t_0$ as the instant of the first image, we denoted, $t$ $= t_0 + k \cdot \Delta t$ with $\Delta t = \mathrm{TS}$. Such a matrix representation was chosen in order to correctly arrange each segmentation mask $\eM^k_p$ according to the spatial and temporal parameters of the dynamic acquisitions. The spatial index $p$ of each $\eM^k_p$ indicated the temporal ordering of acquisition between all planes acquired within each acquisition cycle and $k$ indexed the acquisition cycle. Therefore, as formulated in \eqref{rec_temp}, each $\eM^k_p$ was associated to $\eA_p^{k \cdot N_p + p}$. 

\begin{figure}[!t]
\centerline{\includegraphics[width=0.8\columnwidth]{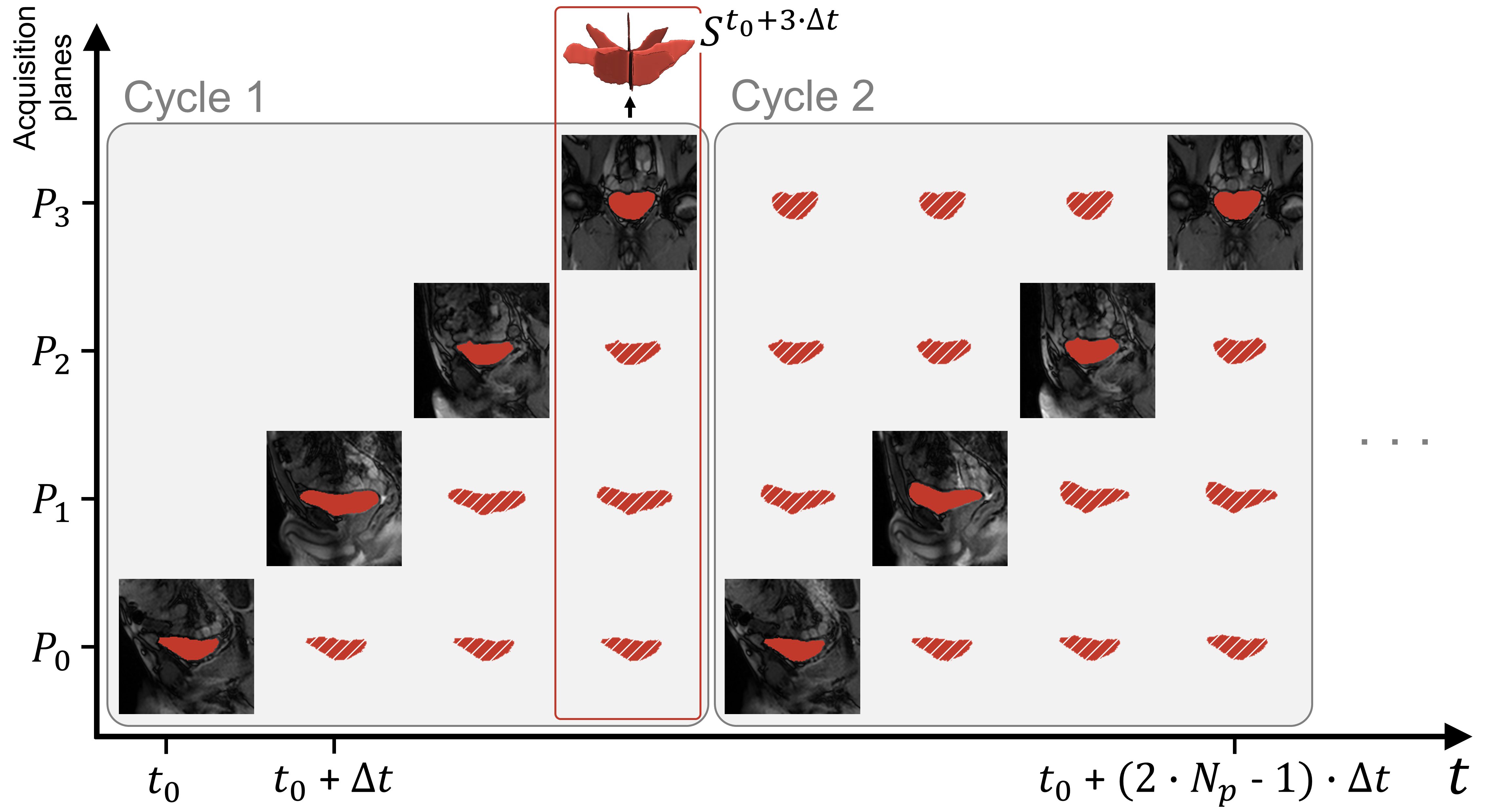}}
\caption{Spatio-temporal configuration of the acquisition planes and temporal reconstruction scheme, illustrated for $C_{star}$. Filled segmentations refer to instants for which an image was acquired. Hatched segmentations correspond to the temporally reconstructed ones. All segmentations of all acquisition planes of a given instant $t$ are combined to provide the skeleton $S^t$ of the organ volume at that time.}
\label{schemaAcq_rec}
\end{figure}

From our 2D acquisition scheme, only one plane of a given configuration $C$ was available at a time $t$, as depicted in Fig.~\ref{schemaAcq_rec}. As a consequence, for a plane position $p$, the segmentation masks had to be interpolated to fulfill the missing acquisition instants in a cycle. The missing segmentations in $\eA_p^t$ were reconstructed by linearly interpolating along the geodesic path between the segmented mask of the previously acquired slice and the next one, as formalized in \eqref{rec_temp}. We advocated such a linear interpolating scheme on the geodesic path since the motion of pelvic organs is assumed to be continuous over the acquisition time of a cycle (less than 1 s). 
\begin{align}
    \begin{split}
        & \eA_p^t=
        \begin{cases}
          \eM_p^{(k//N_p)} & \text{if}\ p = k\tpmod{N_p} \\
          \eM_p^{k'} \circ \phi(x, \frac{k-k'}{N_p}) & \text{otherwise}
        \end{cases} \\
        \hspace{0.5em} with \hspace{1em}& \\
% 		& \eM(p,k') \circ \phi(x,1) \approx \eM(p,k'') \\
		& \eM_p^{k'} \circ \phi(x,1) \approx \eM_p^{k''} \\
		& k' \hspace{0.2em} =  \big( (k-N_p)//N_p \big) \cdot N_p + p \\
		& k'' = k' + N_p
    \end{split}
    \label{rec_temp}
\end{align}

An illustration of the temporal reconstruction approach is given in Fig.~\ref{schemaAcq_rec} with the $C_{star}$ configuration. The scheme is identical whatever the configuration with abscissa values ranged from $0$ to $N_p$ and with $t$ up to $(N_p \cdot N_c - 1)$. 
An horizontal line corresponds to the intermediate masks generated by interpolation between two actually segmented masks.

%%%%%%%%%%%%%%%%%%%%%%%%%%%%%%%%%%%%%%%%%%%%%%%%%%%%%%%%%%%%%%%%%%%%%%%%%
\subsection{Three-dimensional isotropic volume reconstructions} \label{section_3drec}

The temporal interpolation yields spatial and temporal samples of the considered organ. In Fig.~\ref{schemaAcq_rec}, the aggregation of masks on a vertical line corresponds to the creation of what we called the \textit{skeleton} of a dynamic volume at time $t$. But all the masks belonged to 2D spatial domains of the different acquisition planes. Therefore, the next step towards the 3D reconstruction of the dynamic volumes was to compute the affine transformations required to transfer the skeletons into a single 3D isotropic space.
With all segmentations defined in the same 3D domain, volume skeletons $S$ were generated at each instant $t$ as the union of the segmentations from each acquisition plane at that time:
\begin{equation}
S^t = \bigcup_{p=0}^{N_p} {\eA_p}^t \hspace{2em} \forall t \in \intervalleff{N_p - 1}{N_p \cdot (N_c - 1) + 1}
\label{notation_volume}
\end{equation}
Merging process was not performed for the first and last $N_p-1$ volumes because segmentations were not available for all the acquisition planes at these instants (cf. Fig.~\ref{schemaAcq_rec}).

\begin{figure}[!t]
\centerline{\includegraphics[width=0.9\columnwidth]{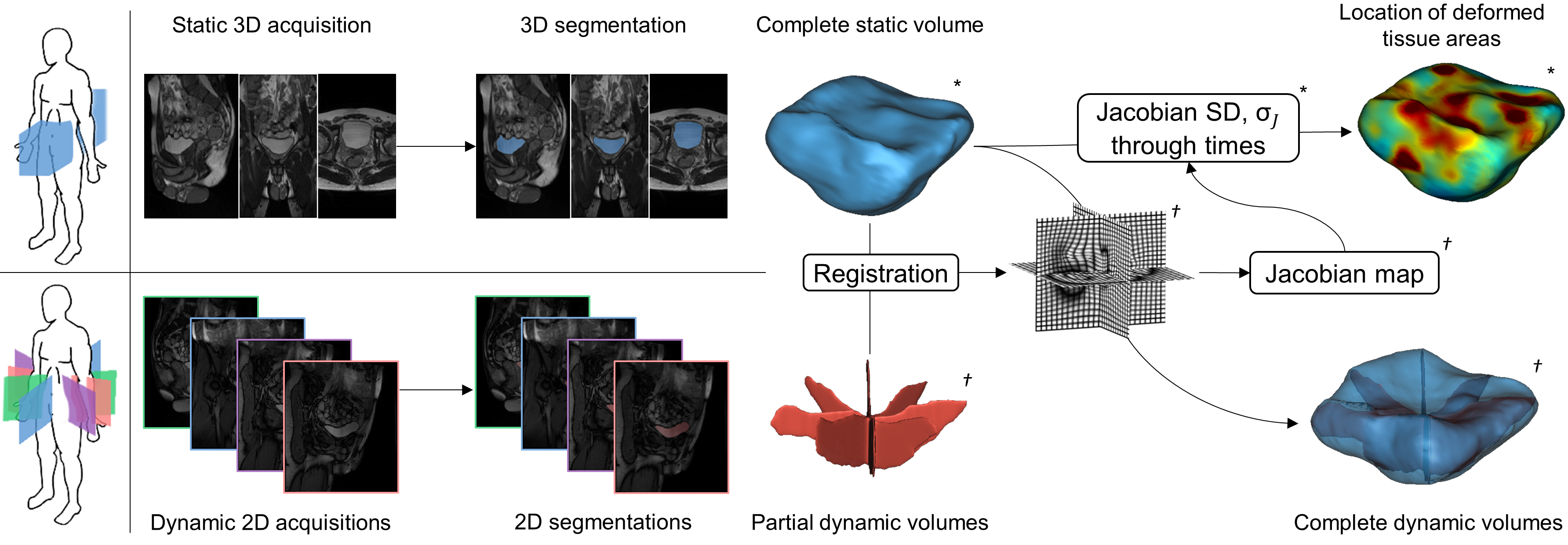}}
\caption{Schematic overview of the reconstruction and characterization pipeline. Reconstruction of the dynamic volumes was achieved by registrating the 3D volume from the static acquisition to the partial volumes from the dynamic acquisitions, obtained after the temporal reconstruction process. The analysis of the resulting deformation fields over time provided the localization of the deformed tissue areas. * static entities, $^\dagger$ dynamic entities.}
\label{pipeline}
\end{figure}

As $S^t$ are partial volumes, the final step towards the 3D reconstruction of the dynamic volumes was to complete these skeletons. To achieve the most natural and realistic reconstruction possible, a subject-specific approach was used. 
The 3D volume resulting from the segmentation of the finely resolved static acquisition was used as a template to be registered to each partial volume (cf. Fig.~\ref{pipeline}). The registration process involved two distinct steps to allow independent analysis of two significant biomechanical mechanisms in the study of pelvic organ disorders: organ displacement and organ deformation. First, organ displacement was estimated at each time by aligning the mass centers of the static 3D volume and each dynamic partial volume. Then, a 3D nonlinear registration was performed between the pre‐aligned volume contours. Nonlinear diffeomorphic model was motivated by the need to register organs that might show dynamics with large deformations. To solve the many-to-few problem of matching complete to partial point clouds, a point/landmark-based similarity metric based on probabilistic estimate was used.
Based on a bi-directional expected matching term between structures to be registered, the PSE metric introduced by Pluta \emph{et al.} \cite{pluta2009appearance} minimizes the distance between each point of a given point-set and the expected corresponding point from the other point-set. Let \{V\} be the set of points describing the contour of the complete static volume $V_s$ and \{R\} the set of points describing the contour of a dynamic volume skeleton. Cardinalities of \{V\} and \{R\} were defined as $m$ and $r$, respectively. Note that $r << m$, as \{R\} described the contour of a partially sampled volume. Our registration problem was therefore to minimize the objective function :
\begin{equation}    
    PSE(\{V\},\{R\})=-\frac{1}{r}\sum_{i=1}^{r}\norm{R_i-\frac{1}{m}\sum_{j=1}^{m}G(R_i;V_j,\sigma_v)V_j}^2
\label{PSE_eq}
\end{equation} 
with $G(R_i;V_j,\sigma_v)$ a normalized Gaussian with mean $V_j$ and standard deviation $\sigma_v$. The standard deviation $\sigma_v$ was empirically set to $1 mm$. Implementation of the registration was performed using the B‐Spline SyN process of the ANTs library \cite{tustison2013explicit}.
Registration processes produced 3D displacement vector fields $u^t = (u_x^t, u_y^t, u_z^t)$, for each instant $t$, enabling to map the complete static volume $V_s$ to each spatially undersampled dynamic volume $S^t$ through the transformations $h^t$ = $v + u^t(v)$ with $v$ a voxel position set by $v = (v_x, v_y, v_z)$. Hence, for $t \in \intervalleff{N_p - 1}{N_p \cdot (N_c - 1) + 1}$, every complete dynamic volumes were generated using $V_s\left(v+u^t(v)\right)$ for each $v$.

%%%%%%%%%%%%%%%%%%%%%%%%%%%%%%%%%%%%%%%%%%%%%%%%%%%%%%%%%%%%%%%%%%%%%%%%%
\subsection{Displacement and deformation characterization}

The radiological interpretation of resultant 4D images can be tedious, thus a quantitative synthesis was sought to facilitate and standardize diagnosis. Taking advantage of the reconstruction process, two quantitative metrics may be directly extracted. The organ displacement recovered from initial affine transformation could be a simple way to characterize organ movement. This metric remained crucial because clinical evaluation of the pelvic dynamics, and hence, the assessment of the pelvic floor disorders is routinely based on the observation of such anatomical landmarks \cite{kobi2018practical}. Similar metrics have already been studied with classic 2D dynamic MRI \cite{rahim2013diffeomorphic,nekooeimehr2018automated}. But to exploit the full potential of the three-dimensional aspect of the reconstruction, 3D organ deformations were analysed using the deformation fields resulting from the non-linear registration.

Jacobian maps were computed to characterize 3D deformations. Let us denote the Jacobian matrix of $h$ as $Dh$ (with the $(i,j)$-$th$ element $\delta h_i / \delta v_j$). The Jacobian map, defined as the determinant of the Jacobian matrix $|Dh(v)|$, encoded the percentages of the dynamic local volume variations at each instant with respect to the static volume image (\emph{i.e.} a value of 0.9 denotes 10\% tissue loss, whereas 1.1 a 10\% tissue increase). The quantity of deformations at each voxel of the organ for each instant $t$ was therefore assessed through Jacobian map $J^t(v) = \mathrm{det} \left(\delta h^t(v) / \delta v\right)$. The projection of each Jacobian map onto the surface of the static volume provided a visualization of the deformations undergone by the organ at the different phases of the loading exercises in a common domain. 
Information provided by all Jacobian maps has also been synthesized into a single map to facilitate the visualization of the deformation areas with a single representation. 
For this purpose, the $\sigma_{J}$ map was calculated for which each voxel $v$ represented the standard deviation (SD) of the values of each $J^t$ map estimated across all $t$ at $v$. Calculation was as follow:
\begin{equation}
\sigma_{J}(v) = \sqrt{\frac{1}{n}\sum_{t=0}^{n}\left(J^t(v) - \overline{J(v)}\right)^2} 
\label{std_jacoco}
\end{equation}
with $n = N_c \cdot N_p - 2(N_p - 1)$, the number of dynamic volumes reconstructed. The $\sigma_{J}$ map projected on a mesh of the reference static volume provided a high-level representation of the location of the most deformed tissue area during a dynamic acquisition sequence.

%%%%%%%%%%%%%%%%%%%%%%%%%%%%%%%%%%%%%%%%%%%%%%%%%%%%%%%%%%%%%%%%%%%%%%%%%
% Results
%%%%%%%%%%%%%%%%%%%%%%%%%%%%%%%%%%%%%%%%%%%%%%%%%%%%%%%%%%%%%%%%%%%%%%%%%
\section{Results}

%%%%%%%%%%%%%%%%%%%%%%%%%%%%%%%%%%%%%%%%%%%%%%%%%%%%%%%%%%%%%%%%%%%%%%%%%
\subsection{Datasets}
Pelvis areas of eight healthy participants (five women) were imaged. Subjects were 27.6 $\pm$ 3.8 years old and weighted 69.3 $\pm$ 11.0 kg. The static acquisition was recorded first during a maximum expiration apnea of 18 seconds. Afterwards, the three dynamic sequences in multi-planar configurations were acquired successively during forced breathing exercises of 1:20 minute each with a short rest between sequences. During these exercises, after 10 seconds of free breathing, the subject alternately inspired and expired at maximum capacity. 
Subjects were also instructed to increase the pelvic pressure to the maximum inspiration and conversely to contract the pelvic floor during the expiration. These actions increased the intra-abdominal pressure, causing deformities of the pelvic organs. The study was approved by the local human research committee and was conducted in conformity with the Declaration of Helsinki.

Since no extrageneous liquid was injected into pelvic cavities in this study, only the segmentation of the bladder was straightforward and the analysis focused exclusively on this organ. For each subject, bladder volumes were generated at a rate of about $1/\mathrm{TS}$ volumes per second. According to the configurations, 394, 492, and approximately 680 volumes were reconstructed for configurations $C_{Star}$, $C_{Grid}$ and $C_{Lines}$, respectively, at a rate of about 5, 8, and 9 volumes per second. As the inspiration/expiration phases of subjects were not paced, the number of respiratory cycles could fluctuate between successive dynamic acquisitions. Depending on the sequences and subjects, pelvic organs undergone from 4 to 12 inspiration/expiration phases. The average post-processing time for the complete methodology was around 18 min per reconstructed dynamic volume, regardless of the geometrical configuration.

Dynamic configuration $C_{lines}$ has only been studied on 3 subjects due to a lack of resources required to the manual segmentation of the dynamic images needed to initialize the semi-automatic process.

Participants were instructed to go to the toilet 2 hours before the exam, to drink moderately up to 30 minutes before and then to fast so that the bladder would not fill during the exam. This way, organ volume did not vary between the different MRI sequences. Bladder volume of each subject was computed from the manual segmentation of the static acquisition. Subjects presented heterogeneous values ranged from 48 cm$^3$ to 403 cm$^3$.

%%%%%%%%%%%%%%%%%%%%%%%%%%%%%%%%%%%%%%%%%%%%%%%%%%%%%%%%%%%%%%%%%%%%%%%%%

\subsection{Spatial discrepancy between segmentations from secant acquisition planes}

Dynamic 2D segmentations were independently realized on differently oriented planes. As only one plane could be imaged at a time, the segmentations from these planes naturally had a time offset between them (cf. Fig.~\ref{schemaAcq_rec}). The evaluation of the discrepancy between segmentations from secant acquisition planes aimed to validate the relevance of the temporal reconstruction process presented in section \ref{section_tempRec} for the reduction of this temporal bias.

\begin{figure}[!t]
\centerline{\includegraphics[width=0.9\columnwidth]{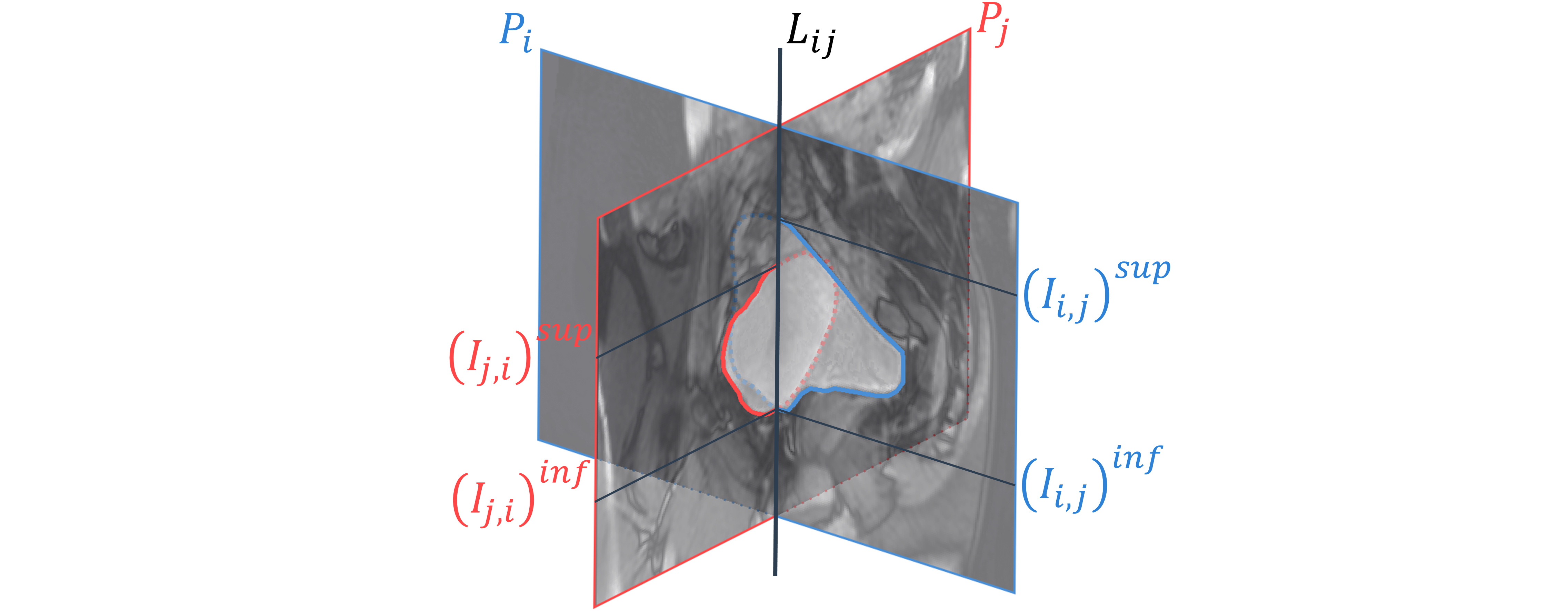}}
\caption{Schematic illustration of the spatial inconsistency between two segmentations from two secant acquisition planes.}
\label{recTemporal_schema}
\end{figure}

Let $L_{ij}$ be the line defined as the intersection of two secant acquisition planes: $L_{ij} = \eP_i \cap \eP_j$. The intersection of $L_{ij}$ with the segmentation mask $\eA_i^t$ related to the plane $\eP_i$ provided the set of points $I_{i,j}^t = \{ L_{ij} \cap \eA_i^t \}$. From this set, the inferior and the superior intersection points were denoted as $(I_{i,j}^t)^{sup}$ and $(I_{i,j}^t)^{inf}$, respectively. 
A schematic illustration is depicted in Fig.~\ref{recTemporal_schema}. In this figure, for visual reasons, the two secant planes were taken at very different times, each corresponding to one of the inspiration/expiration phases. In practice, during this study, distances between segmentation intersections remained narrower thanks to the high speed of plane acquisitions.

Spatial discrepancy $\zeta$ between two segmentations from two secant planes was defined as the sum of the two Euclidean distances between the superior and inferior intersection points:
\begin{align} \begin{split}
    \zeta(\eA_i,\eA_j)^t = \norm{(I_{i,j}^t)^{sup} - (I_{j,i}^t)^{sup}} \\
    + \norm{(I_{i,j}^t)^{inf} - (I_{j,i}^t)^{inf}}
\end{split} \label{dist_planes} \end{align}

Spatial discrepancy was calculated between each pair of secant planes for each configuration except for $C_{lines}$ (which was only composed of parallel planes). Spatial discrepancy scores for all subjects and all instants were 1.71 $\pm$ 1.24 mm for $C_{star}$ and 2.51 $\pm$ 1.57 mm for $C_{grid}$, which corresponds to only a few pixels. 

%%%%%%%%%%%%%%%%%%%%%%%%%%%%%%%%%%%%%%%%%%%%%%%%%%%%%%%%%%%%%%%%%%%%%%%%%
\subsection{Consistency of three-dimensional reconstructions}

The B‐spline SyN registration implemented for the 3D reconstruction process was optimized as follow: the gradient step was 0.1 for each of the four multi-resolution levels with shrink factors of \{$6, 4, 2, 1$\} and no smoothing sigmas. The number of iterations per level were \{$250, 200, 150, 100$\} with a stepping out criterion based on a convergence threshold of $10^{-6}$ within 15 iterations. Knot spacing was set to $44$ mm for the B-spline smoothing on the update displacement field at the base level which is reduced by a factor of two for each successive multiresolution level. This yielded a final knot spacing of $5.5$ mm. Such low regularization scheme was retained because pelvic organ deformations are mostly global and no high frequency deformations had to be permitted in the registration process. Furthermore, this optimization preserved the overall topology of the template static organ. No restrictions were applied to the smoothing of the total displacement field as this did not improve the results.

\begin{table}[!h]
\centering
\caption{Evaluation of three-dimensional reconstructions}
%\small
\setlength{\tabcolsep}{3pt}
\begin{tabular}{|p{70pt}|p{45pt}|c|c|c|}
\hline
\rowcolor{Gray}
Configuration & Subject & RAVD (\%) & HD (mm) & MD (mm) \\ \hline
$C_{star}$  & \#1       & 1.64 $\pm$ 1.47   & 1.37 $\pm$ 0.24   & 0.21 $\pm$ 0.06   \\
\rowcolor{Gray}
            & \#2       & 1.88 $\pm$ 1.80   & 1.41 $\pm$ 0.28   & 0.18 $\pm$ 0.03   \\
            & \#3       & 1.95 $\pm$ 1.41   & 1.36 $\pm$ 0.27   & 0.17 $\pm$ 0.04   \\
\rowcolor{Gray}
            & \#4       & 2.82 $\pm$ 1.61   & 1.28 $\pm$ 0.22   & 0.18 $\pm$ 0.03   \\
            & \#5       & 3.40 $\pm$ 0.94   & 1.33 $\pm$ 0.23   & 0.17 $\pm$ 0.02   \\
\rowcolor{Gray}
            & \#6       & 1.66 $\pm$ 0.67   & 1.31 $\pm$ 0.23   & 0.16 $\pm$ 0.02   \\
            & \#7       & 5.94 $\pm$ 3.31   & 1.32 $\pm$ 0.23   & 0.17 $\pm$ 0.02   \\
\rowcolor{Gray}
            & \#8       & 1.58 $\pm$ 1.23   & 1.42 $\pm$ 0.29   & 0.19 $\pm$ 0.08   \\
& \textbf{Overall} & \textbf{2.61 $\pm$ 2.22} & \textbf{1.35 $\pm$ 0.25} & \textbf{0.18 $\pm$ 0.04}  \\ \hline

\rowcolor{Gray}
$C_{grid}$  & \#1       & 2.70 $\pm$ 2.49   & 2.44 $\pm$ 0.68   & 0.45 $\pm$ 0.17   \\
            & \#2       & 2.36 $\pm$ 2.12   & 2.07 $\pm$ 0.33   & 0.33 $\pm$ 0.09   \\
\rowcolor{Gray}
            & \#3       & 2.63 $\pm$ 1.68   & 2.86 $\pm$ 0.63   & 0.65 $\pm$ 0.15   \\
            & \#4       & 2.74 $\pm$ 2.13   & 1.76 $\pm$ 0.29   & 0.28 $\pm$ 0.07   \\
\rowcolor{Gray}
            & \#5       & 1.26 $\pm$ 0.67   & 1.65 $\pm$ 0.28   & 0.23 $\pm$ 0.05   \\
            & \#6       & 1.67 $\pm$ 1.03   & 1.82 $\pm$ 0.23   & 0.25 $\pm$ 0.05   \\
\rowcolor{Gray}
            & \#7       & 2.00 $\pm$ 1.46   & 1.76 $\pm$ 0.35   & 0.25 $\pm$ 0.05   \\
            & \#8       & 1.51 $\pm$ 1.21   & 1.75 $\pm$ 0.37   & 0.22 $\pm$ 0.09   \\
\rowcolor{Gray}
& \textbf{Overall} & \textbf{2.11 $\pm$ 1.79} & \textbf{2.01 $\pm$ 0.58} & \textbf{0.33 $\pm$ 0.17}  \\ \hline
            
$C_{lines}$ & \#5       & 1.40 $\pm$ 0.91   & 3.60 $\pm$ 0.70   & 0.85 $\pm$ 0.15   \\
\rowcolor{Gray}
            & \#6       & 1.03 $\pm$ 0.70   & 4.49 $\pm$ 0.89   & 1.07 $\pm$ 0.20   \\
            & \#7       & 5.28 $\pm$ 2.85   & 3.31 $\pm$ 0.60   & 0.75 $\pm$ 0.14   \\
\rowcolor{Gray}
& \textbf{Overall} & \textbf{2.57 $\pm$ 2.62} & \textbf{3.80 $\pm$ 0.90} & \textbf{0.89 $\pm$ 0.21} \\ \hline
% \multicolumn{2}{|c|}{Global Average} & 2.39 $\pm$ 2.16   & 2.19 $\pm$ 1.09   & 0.41 $\pm$ 0.31\\ \hline
\multicolumn{5}{p{350pt}}{RVD = relative volume deviation, HD = Hausdorff distance, MD = mean distance. Values are mean $\pm$ standard deviation of all reconstructed volumes.}\\
\end{tabular}
\label{revConsistency}
\end{table}

Validation of the 3D organ reconstructions was a difficult process due to the lack of ground truth as a point of comparison. Therefore, the evaluation focused on the consistency of the reconstructed volumes through two metrics: accuracy of the registration and conservation of the bladder volume.
As the bladder is an incompressible organ (filled with a liquid), its volume has to remain unchanged during the registration process when the static volume is deformed to match the shape of the partial dynamic volumes. Relative absolute volume deviation (RAVD) was computed at each instant between the reference volume and the reconstructed volume. The time averages of the values are gathered in Table \ref{revConsistency}. Average volume deviations were quite similar for all configurations with a mean value of $2.39\% \pm 2.16\%$. Outliers in Subject 7's data might be attributed to the fact that the subject did not comply with pre-acquisition fluid consumption guidelines and therefore had the lowest bladder volume among all subjects. 

Evaluation of the mapping of the deformed static volume onto the organ volume skeletons provided the assessment of the registration accuracy. For each point of the contour of the dynamic volume skeleton, the distance to the nearest point of the contour of the reconstructed volume was calculated. From these distances were extracted the maximum value and the average value, described as the non-symmetric Hausdorff distance (HD) and the mean distance (MD), respectively (Table \ref{revConsistency}). The average HD and MD over time reflected great registration accuracy for all configurations with mean values of 2.19 $\pm$ 2.09 and 0.41 $\pm$ 0.31 mm, respectively.
It should be noted that comparison of the results between configurations must take into consideration that HD and MD were computed in the discrete 3D domains of the reconstructions and were therefore directly related to their isotropic spatial resolution. As reported in Table \ref{paramMRIdyn}, voxel size was 1.09 mm for $C_{star}$, 1.36 mm for $C_{grid}$ and 1.82 mm for $C_{lines}$. These results once expressed in voxels are quite similar between configurations and accuracy is about 1 or 2 voxels overall.

%%%%%%%%%%%%%%%%%%%%%%%%%%%%%%%%%%%%%%%%%%%%%%%%%%%%%%%%%%%%%%%%%%%%%%%%%

\begin{figure}[!b]
\centerline{\includegraphics[width=0.9\columnwidth]{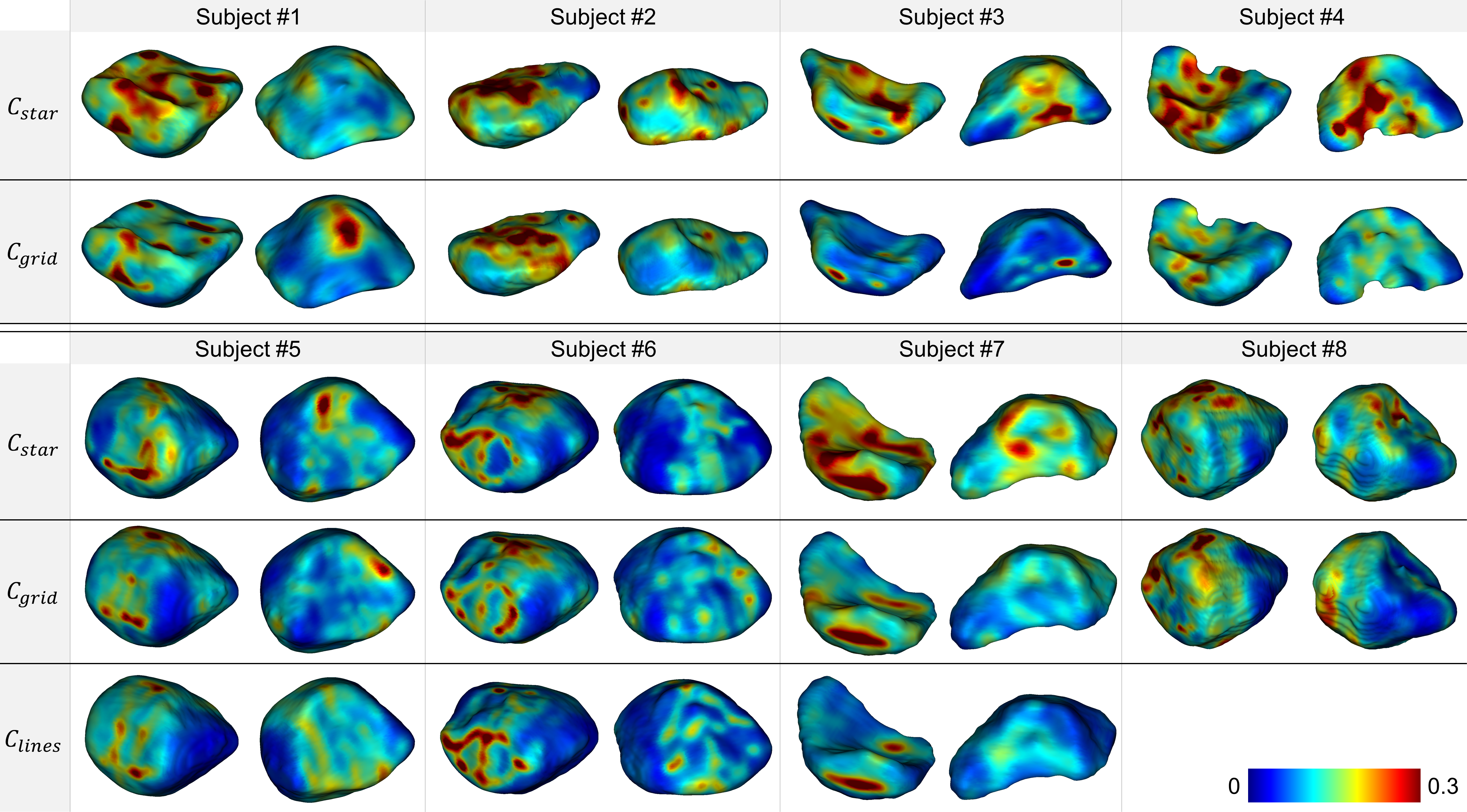}}
\caption{Anterior (left) and posterior (right) views for each subject and configuration of the $\sigma_{J}$ map projected on the corresponding reference static volumes. Reddest areas indicate the location of the highest strain areas on the organ surfaces during dynamic acquisition sequences.}
\label{jacoco}
\end{figure}

%%%%%%%%%%%%%%%%%%%%%%%%%%%%%%%%%%%%%%%%%%%%%%%%%%%%%%%%%%%%%%%%%%%%%%%%%
\subsection{High-level representation of pelvic organ deformation zones}

Reconstructed volumes over skeletons and projections of the associated Jacobian maps are available for two subjects as supporting materials. Projections of the $\sigma_{J}$ maps on meshes of the static reference volumes are shown in Fig.~\ref{jacoco} for each subject and each configuration. Visualizations were performed using the Anatomist software \cite{riviere2000structural}. 
At each point, the value for the intersection between the mesh of the static reference volume and the $\sigma_{J}$ map, was estimated as the mean of the values within a sphere of 4 mm radius, almost 3 voxels, around the intersection point. This projection scheme was common to all configurations to provide a common reference for comparison regardless the resolutions of the reconstructed volumes.
The reddest areas indicate the locations that underwent the largest deformations during the loading exercises.
Note that volume scales are not the same for each subject in Fig.~\ref{jacoco}, \emph{e.g.} bladder volume was 48 cm$^3$ for subject 7 whereas it was 403 cm$^3$ for subject 8.

Forced breathing exercises with maximum contraction are hardly reproducible for untrained subjects. Deformation magnitudes were different for each dynamic sequence and consequently differed for each geometric configuration $C$. 
For instance, in Fig.~\ref{jacoco}, subjects 4 and 5 did not exert the same intensities during each loading exercise explaining the main scale differences. On the whole, no visual correlation can be done between geometry configuration and amplitude of the deformations. Dynamic sequences with the highest deformation characteristics are $C_{star}$ for subjects 1, 2, 7, and 8, $C_{grid}$ for subjects 3, 4, and 5, and $C_{lines}$ for subject 6. 
Intra-subject comparison of deformation amplitudes might not be relevant and was not the focus of our study. However, regardless of the load magnitude applied to the bladder, the intra-subject locations of the most deformed tissue areas are quite similar between dynamic sequences. Overall, tissue areas that have undergone deformities are located on the anterior and superior parts of bladders due to the abdominal viscera pushing against the bladder during loading exercises. Few deformation zones are visible on the inferior part due to a strong pelvic floor in our young healthy volunteers. However, inferior deformations in patients could indicate pelvic floor weaknesses.

%%%%%%%%%%%%%%%%%%%%%%%%%%%%%%%%%%%%%%%%%%%%%%%%%%%%%%%%%%%%%%%%%%%%%%%%%
% Discussion
%%%%%%%%%%%%%%%%%%%%%%%%%%%%%%%%%%%%%%%%%%%%%%%%%%%%%%%%%%%%%%%%%%%%%%%%%
\section{Discussion}

While extensive research has been conducted on the movement of organs such as the brain \cite{bauer2013survey} and heart \cite{wang2011cardiac}, little work has been done on the analysis of the pelvic organ dynamics. 
Most of the studies have focused either on the realization of biomechanical simulators generated from 3D static acquisition \cite{chen2015female}, or on the characterization of organ deformations based on a single 2D dynamic slice \cite{rahim2013diffeomorphic, nekooeimehr2018automated}.
Recently, Courtecuisse \emph{et al.} \cite{courtecuisse2020three} has proposed a registration approach to deform pelvic organs acquired in 3D static acquisition using single-plane 2D dynamic acquisition and \emph{a priori} knowledge of boundary conditions and mechanical laws of the pelvic organs. This biomechanical knowledge had to be injected into the registration model because the ground truth information from a single 2D section is not enough to accurately guide organs that deform nonlinearly in all 3 dimensions. This approach remains limited because it depends on pelvic tissue characterization, which remains unclear with high inter-subject variability, and the realization of patient-specific models would require invasive procedures to probe the biomechanical properties \emph{in vivo}.
In our study, we provided 3D spatial coverage of the organs through the acquisition of several dynamic slices. 
Dynamic volumes did not require to be simulated in a biomechanical sense and were reconstructed independently of any mechanical model. And the subject-specific initial geometric model required for the 3D reconstruction process was readily available from a fast 3D static acquisition. Two possible static states were available for each subject from maximum inspiration or maximum expiration breath-hold. 
The latter one was retained because it was the state closest to the natural condition of organs and, in pathological routines, will be less painful than inspiration. 
In further studies with patient recruitment, the deformation of the static model on dynamic volumes may require refinement to account for pelvic prolapse. This study emphasized metrics to assess the adequacy of bladder reconstructions. Using these metrics, we demonstrated the appropriateness of the registration method for the reconstruction by preserving an average RAVD value around $2.5\%$ and respecting the ground-truth contours provided by the dynamic multi-slices with average HD values of 2.2 $\pm$ 1.1 mm and average MD values of 0.4 $\pm$ 0.3 mm.

%The full 3D spatial organ coverage imposed cautious multi-plane acquisitions. As the planes could not be imaged simultaneously there were temporal gaps between frame acquisitions that could result in spatial inconsistencies between the associated segmentations. Eventually, we proposed a temporal reconstruction scheme which successfully limited the spatial gap $\zeta$ between segmentations within the range of a few voxels.%
%One key process of our study is the segmentation stage. Given the state of the art, it would have been too ambitious to consider a single, fully automatic method for segmenting the pelvic organs along various frame orientations.%

One limitation is the semi-automated segmentation that requires a minimum user interaction. For pelvic organs, existing automatic segmentation methods are generally dedicated to a specific type of datasets, either static 3D images or 2D axial or sagittal images. Most of approaches are based on deformable models \cite{namias2014uterus} which are rather difficult to parameterize even for one type of images. So, this step remained the only one that has not been fully automated in our pipeline. Still, one of the originalities of our study was to propose dynamic acquisitions under different planes with open 3D geometric configurations. 
The advantages of the semi-automatic method that we propose are manifold: First, the time spent to perform manual segmentation has been reduced by about $90\%$ as only one in ten slices was manually segmented; Second, the proposed method is suitable on all planes, regardless of their orientations, resolutions, and contrasts. 
Thus, the proposed method adjusts to all these acquisition parameters by propagating organ contours that were manually initialized on a few instants, from one slice to another similar one.
In further studies, such semi-automatic method of segmentation will facilitate the creation of labeled datasets of 2D dynamic MRI and hence will allow to investigate the relevance of deep-learning methods for pelvic organ segmentation. To date, these approaches have been scarcely used and still provide insufficient results, with average DSC around $85\%$ \cite{feng2020convolutional}. 

Within one cycle, planes are not acquired simultaneously, thus skeleton definition is ill-posed due to the spatial discrepancy $\zeta$ between crossing planes segmentations. This element was alleviated by proposed temporal interpolation that joined virtually simultaneous crossing planes.

Dynamic volumes from each configuration could not be directly compared because subjects could not reproduce same pelvic organ movements between each dynamic acquisition. Frequency of pushing phases and magnitude of the load applied to the organs were highly variable. Volumes, and associated $J$ maps, could not be compared one to one from one configuration to another. However, the higher-level $\sigma_{J}$ maps allowed to compare the spatial distribution of deformed tissue areas. For all subjects, the spatial distribution of deformed tissue regions was similar between configurations. 
Although $\sigma_{J}$ map has no direct physical meaning, its projection on the reference volume provided a preliminary mechanical characterization of the organs.

Overall, our processing pipeline proved to be compatible with all three configurations in terms of bladder reconstruction and deformation characterization, with equivalent end results. One of the three dynamic acquisition planes configurations can be chosen to ensure either complete spatial coverage ($C_{lines}$) or high spatial resolution ($C_{star}$), and combined with our methodology to provide dynamic 3D representation of the bladder during loading exercises. The proposed pipeline is directly applicable in clinics and can be added to a routine examination as the acquisition time required is minimal and sequences are standard. In practice, only a 20 seconds static apnea sequence and a multi-planar acquisition of one straining exercise are necessary.

% %%%%%%%%%%%%%%%%%%%%%%%%%%%%%%%%%%%%%%%%%%%%%%%%%%%%%%%%%%%%%%%%%%%%%%%%%
% % Conclusion
% %%%%%%%%%%%%%%%%%%%%%%%%%%%%%%%%%%%%%%%%%%%%%%%%%%%%%%%%%%%%%%%%%%%%%%%%%
% \section{Conclusion}

We have designed a complete methodology for the 3D+t representation of pelvic organs, during loading exercises, directly combined with a high-level 3D representation of the most strained areas on the organ surfaces. To our knowledge, this study is the first to propose a dynamic 3D real-time observation of the pelvic region as well as a 3D representation of the organs deformations. Our proposal opens novel perspectives towards real-time dynamic 3D pelvic imaging and biomechanical pelvic organs characterization. % \cite{courtecuisse2020three}.

In this study, three different geometrical configurations of dynamic multi-planar acquisitions have been proposed to provide the first dynamic 3D representation of pelvic organs. Two of them were particularly original, $C_{star}$ and $C_{grid}$, because they were composed of planes oriented differently from the sagittal plane as usually imagined in clinics. In this study, we verified that all three configurations could provide deformation characterization. In consequence, the choice of a multi-planar configuration depends on the preference for spatial coverage or spatial resolution. In clinical practice, it might be easier to configure $C_{lines}$, but configuring $C_{star}$ should be straight-forward and enable improved visualization of smaller structures. %One of our aims was to investigate whether one configuration was more advantageous than another. However, regardless of the configuration used, all of the processes leading to the dynamic 3D representation of the organs performed identically, providing similar validation results.

To allow a direct transfer to clinics, only the acceleration methods available on the clinical machines were considered. However, the recently developed simultaneous multi‐slice acquisition would allow the simultaneous acquisition of several parallel planes at the same time \cite{rapacchi2019simultaneous}.
Besides, future studies will be conducted with the ethical authorization to inject extrageneous liquid into the rectal and vaginal cavities in order to segment all pelvic organs and study the complete pelvic dynamics.
The ability to now obtain dynamic 3D volumes allows the development of more advanced 3D shape descriptors \cite{makki2021new} to better characterize the degree of severity of a pathology for diagnostic assistance. These studies are a 3D extension of descriptors such as we already proposed for 2D dynamic pelvic acquisitions \cite{rahim2013diffeomorphic}. All this leads to the eventuality of a 3D ground truth that will allow the validation of future biomechanical models at least.

%%%%%%%%%%%%%%%%%%%%%%%%%%%%%%%%%%%%%%%%%%%%%%%%%%%%%%%%%%%%%%%%%%%%%%%%%
% Acknowledgements 
%%%%%%%%%%%%%%%%%%%%%%%%%%%%%%%%%%%%%%%%%%%%%%%%%%%%%%%%%%%%%%%%%%%%%%%%%
\section{Acknowledgements}

This study was ethically approved by the following institutional review board: “comité de protection des personnes Sud-Ouest et Outre-Mer 1”, from September 21st 2020, ID RCB : 2020-A00673-36. This research did not receive any specific grant from funding agencies in the public, commercial, or not-for-profit sectors.

%%%%%%%%%%%%%%%%%%%%%%%%%%%%%%%%%%%%%%%%%%%%%%%%%%%%%%%%%%%%%%%%%%%%%%%%%
\bibliography{mybibfile}

\begin{thebibliography}{10}
\expandafter\ifx\csname url\endcsname\relax
  \def\url#1{\texttt{#1}}\fi
\expandafter\ifx\csname urlprefix\endcsname\relax\def\urlprefix{URL }\fi
\expandafter\ifx\csname href\endcsname\relax
  \def\href#1#2{#2} \def\path#1{#1}\fi

\bibitem{nygaard2008prevalence}
I.~Nygaard, M.~D. Barber, K.~L. Burgio, K.~Kenton, S.~Meikle, J.~Schaffer,
  C.~Spino, W.~E. Whitehead, J.~Wu, D.~J. Brody, et~al., Prevalence of
  symptomatic pelvic floor disorders in us women, Jama 300~(11) (2008)
  1311--1316.

\bibitem{wu2014prevalence}
J.~M. Wu, C.~P. Vaughan, P.~S. Goode, D.~T. Redden, K.~L. Burgio, H.~E.
  Richter, A.~D. Markland, Prevalence and trends of symptomatic pelvic floor
  disorders in us women, Obstetrics and gynecology 123~(1) (2014) 141.

\bibitem{delancey2005hidden}
J.~O. DeLancey, The hidden epidemic of pelvic floor dysfunction: achievable
  goals for improved prevention and treatment, American journal of obstetrics
  and gynecology 192~(5) (2005) 1488--1495.

\bibitem{gurland2021consensus}
B.~H. Gurland, G.~Khatri, R.~Ram, T.~L. Hull, E.~Kocjancic, L.~H. Quiroz, R.~F.
  El~Sayed, K.~R. Jambhekar, V.~Chernyak, R.~M. Paspulati, et~al., Consensus
  definitions and interpretation templates for magnetic resonance imaging of
  defecatory pelvic floor disorders, International Urogynecology Journal
  32~(10) (2021) 2561--2574.

\bibitem{woodfield2010imaging}
C.~A. Woodfield, S.~Krishnamoorthy, B.~S. Hampton, J.~M. Brody, Imaging pelvic
  floor disorders: trend toward comprehensive mri, American Journal of
  Roentgenology 194 (2010) 1640--1649.

\bibitem{el2017magnetic}
R.~F. El~Sayed, C.~D. Alt, F.~Maccioni, M.~Meissnitzer, G.~Masselli,
  L.~Manganaro, V.~Vinci, D.~Weishaupt, ESUR, E.~P. F.~W. Group, et~al.,
  Magnetic resonance imaging of pelvic floor dysfunction-joint recommendations
  of the esur and esgar pelvic floor working group, European radiology 27~(5)
  (2017) 2067--2085.

\bibitem{zhang2022dynamic}
H.~Zhang, Z.~Wang, X.~Xiao, J.~Wang, B.~Zhou, Dynamic magnetic resonance
  imaging evaluation before and after operation for pelvic organ prolapse,
  Abdominal Radiology 47~(2) (2022) 848--857.

\bibitem{rahim2013diffeomorphic}
M.~Rahim, M.-E. Bellemare, R.~Bulot, N.~Pirr{\'o}, A diffeomorphic mapping
  based characterization of temporal sequences: application to the pelvic organ
  dynamics assessment, Journal of mathematical imaging and vision 47~(1-2)
  (2013) 151--164.

\bibitem{nekooeimehr2018automated}
I.~Nekooeimehr, S.~K. Lai-Yuen, P.~Bao, A.~Weitzenfeld, S.~Hart, Automated
  contour tracking and trajectory classification of pelvic organs on dynamic
  mri, Journal of Medical Imaging 5~(1) (2018) 014008.

\bibitem{salsi2017three}
G.~Salsi, I.~Cataneo, G.~Dodaro, N.~Rizzo, G.~Pilu, M.~S. Gasc{\`o}n,
  A.~Youssef, Three-dimensional/four-dimensional transperineal ultrasound:
  clinical utility and future prospects, International journal of women's
  health 9 (2017) 643.

\bibitem{weber2005pelvic}
A.~M. Weber, H.~E. Richter, Pelvic organ prolapse, Obstetrics \& Gynecology
  106~(3) (2005) 615--634.

\bibitem{maher2013surgical}
C.~Maher, B.~Feiner, K.~Baessler, C.~Schmid, Surgical management of pelvic
  organ prolapse in women, Cochrane database of systematic reviews~(4).

\bibitem{abed2011incidence}
H.~Abed, D.~D. Rahn, L.~Lowenstein, E.~M. Balk, J.~L. Clemons, R.~G. Rogers,
  S.~R.~G. of~the Society~of Gynecologic~Surgeons, et~al., Incidence and
  management of graft erosion, wound granulation, and dyspareunia following
  vaginal prolapse repair with graft materials: a systematic review,
  International urogynecology journal 22~(7) (2011) 789--798.

\bibitem{bellemare2007toward}
M.-E. Bellemare, N.~Pirro, L.~Marsac, O.~Durieux, Toward the simulation of the
  strain of female pelvic organs, in: 2007 29th Annual International Conference
  of the IEEE Engineering in Medicine and Biology Society, IEEE, 2007, pp.
  2752--2755.

\bibitem{chen2015female}
Z.-W. Chen, P.~Joli, Z.-Q. Feng, M.~Rahim, N.~Pirr{\'o}, M.-E. Bellemare,
  Female patient-specific finite element modeling of pelvic organ prolapse
  (pop), Journal of biomechanics 48~(2) (2015) 238--245.

\bibitem{rubod2012biomechanical}
C.~Rubod, M.~Brieu, M.~Cosson, G.~Rivaux, J.-C. Clay, L.~de~Landsheere,
  B.~Gabriel, Biomechanical properties of human pelvic organs, Urology 79~(4)
  (2012) 968--e17.

\bibitem{rahim2011quantiative}
M.~Rahim, M.-E. Bellemare, N.~Pirr{\'o}, R.~Bulot, A quantitative approach for
  the assessment of the pelvic dynamics modeling, IRBM 32~(5) (2011) 311--315.

\bibitem{courtecuisse2020three}
H.~Courtecuisse, Z.~Jiang, O.~Mayeur, J.~Witz, P.~Lecomte-Grosbras, M.~Cosson,
  M.~Brieu, S.~Cotin, Three-dimensional physics-based registration of pelvic
  system using 2d dynamic magnetic resonance imaging slices, Strain (2020)
  e12339.

\bibitem{gordon2019framework}
M.~T. Gordon, J.~O.~L. DeLancey, A.~Renfroe, A.~Battles, L.~Chen, Development
  of anatomically based customizable three-dimensional finite-element model of
  pelvic floor support system: Pop-sim1.0, Interface Focus 9~(4) (2019)
  20190022.
\newblock \href {http://dx.doi.org/10.1098/rsfs.2019.0022}
  {\path{doi:10.1098/rsfs.2019.0022}}.

\bibitem{mcelroy2020combined}
S.~McElroy, G.~Ferrazzi, M.~S. Nazir, K.~P. Kunze, R.~Neji, P.~Speier,
  D.~St{\"a}b, C.~Forman, R.~Razavi, A.~Chiribiri, et~al., Combined
  simultaneous multislice bssfp and compressed sensing for first-pass
  myocardial perfusion at 1.5 t with high spatial resolution and coverage,
  Magnetic resonance in medicine 84~(6) (2020) 3103--3116.

\bibitem{feng2016xd}
L.~Feng, L.~Axel, H.~Chandarana, K.~T. Block, D.~K. Sodickson, R.~Otazo,
  Xd-grasp: golden-angle radial mri with reconstruction of extra motion-state
  dimensions using compressed sensing, Magnetic resonance in medicine 75~(2)
  (2016) 775--788.

\bibitem{ogier20193d}
A.~C. Ogier, S.~Rapacchi, A.~Le~Troter, M.-E. Bellemare, 3d dynamic mri for
  pelvis observation-a first step, in: 2019 IEEE 16th International Symposium
  on Biomedical Imaging (ISBI 2019), IEEE, 2019, pp. 1801--1804.

\bibitem{lustig2007sparse}
M.~Lustig, D.~Donoho, J.~M. Pauly, Sparse mri: The application of compressed
  sensing for rapid mr imaging, Magnetic Resonance in Medicine 58~(6) (2007)
  1182--1195.

\bibitem{breuer2005dynamic}
F.~A. Breuer, P.~Kellman, M.~A. Griswold, P.~M. Jakob, Dynamic autocalibrated
  parallel imaging using temporal grappa (tgrappa), Magnetic Resonance in
  Medicine 53~(4) (2005) 981--985.

\bibitem{beg2005computing}
M.~F. Beg, M.~I. Miller, A.~Trouv{\'e}, L.~Younes, Computing large deformation
  metric mappings via geodesic flows of diffeomorphisms, International journal
  of computer vision 61~(2) (2005) 139--157.

\bibitem{pasquier2007automatic}
D.~Pasquier, T.~Lacornerie, M.~Vermandel, J.~Rousseau, E.~Lartigau,
  N.~Betrouni, Automatic segmentation of pelvic structures from magnetic
  resonance images for prostate cancer radiotherapy, International Journal of
  Radiation Oncology* Biology* Physics 68~(2) (2007) 592--600.

\bibitem{duan2010coupled}
C.~Duan, Z.~Liang, S.~Bao, H.~Zhu, S.~Wang, G.~Zhang, J.~J. Chen, H.~Lu, A
  coupled level set framework for bladder wall segmentation with application to
  mr cystography, IEEE transactions on medical imaging 29~(3) (2010) 903--915.

\bibitem{ma2013segmentation}
Z.~Ma, R.~M.~N. Jorge, T.~Mascarenhas, J.~M.~R. Tavares, Segmentation of female
  pelvic organs in axial magnetic resonance images using coupled geometric
  deformable models, Computers in biology and medicine 43~(4) (2013) 248--258.

\bibitem{namias2014uterus}
R.~Nam{\'\i}as, M.-E. Bellemare, M.~Rahim, N.~Pirr{\'o},
  \href{https://doi.org/10.1117/12.2043617}{Uterus segmentation in dynamic mri
  using lbp texture descriptors}, in: Medical Imaging 2014: Image Processing,
  Vol. 9034, International Society for Optics and Photonics, SPIE, 2014, pp.
  1009 -- 1017.
\newblock \href {http://dx.doi.org/10.1117/12.2043617}
  {\path{doi:10.1117/12.2043617}}.
\newline\urlprefix\url{https://doi.org/10.1117/12.2043617}

\bibitem{jiang2015b}
Z.~Jiang, J.-F. Witz, P.~Lecomte-Grosbras, J.~Dequidt, C.~Duriez, M.~Cosson,
  S.~Cotin, M.~Brieu, B-spline based multi-organ detection in magnetic
  resonance imaging, Strain 51~(3) (2015) 235--247.

\bibitem{nie2018strainet}
D.~Nie, L.~Wang, Y.~Gao, J.~Lian, D.~Shen, Strainet: Spatially varying
  stochastic residual adversarial networks for mri pelvic organ segmentation,
  IEEE transactions on neural networks and learning systems 30~(5) (2018)
  1552--1564.

\bibitem{ogier2017individual}
A.~Ogier, M.~Sdika, A.~Foure, A.~Le~Troter, D.~Bendahan, Individual muscle
  segmentation in mr images: A 3d propagation through 2d non-linear
  registration approaches, in: Engineering in Medicine and Biology Society
  (EMBC), 2017 39th Annual International Conference of the IEEE, IEEE, 2017,
  pp. 317--320.

\bibitem{pluta2009appearance}
J.~Pluta, B.~B. Avants, S.~Glynn, S.~Awate, J.~C. Gee, J.~A. Detre, Appearance
  and incomplete label matching for diffeomorphic template based hippocampus
  segmentation, Hippocampus 19~(6) (2009) 565--571.

\bibitem{tustison2013explicit}
N.~J. Tustison, B.~B. Avants, Explicit b-spline regularization in diffeomorphic
  image registration, Frontiers in neuroinformatics 7 (2013) 39.

\bibitem{kobi2018practical}
M.~Kobi, M.~Flusberg, V.~Paroder, V.~Chernyak, Practical guide to dynamic
  pelvic floor mri, Journal of Magnetic Resonance Imaging 47~(5) (2018)
  1155--1170.

\bibitem{riviere2000structural}
D.~Rivi{\`e}re, D.~Papadopoulos-Orfanos, J.~R{\'e}gis, J.-F. Mangin, A
  structural browser of brain anatomy, NeuroImage 11~(5).

\bibitem{bauer2013survey}
S.~Bauer, R.~Wiest, L.-P. Nolte, M.~Reyes, A survey of mri-based medical image
  analysis for brain tumor studies, Physics in Medicine \& Biology 58~(13)
  (2013) R97.

\bibitem{wang2011cardiac}
H.~Wang, A.~A. Amini, Cardiac motion and deformation recovery from mri: a
  review, IEEE transactions on medical imaging 31~(2) (2011) 487--503.

\bibitem{feng2020convolutional}
F.~Feng, J.~A. Ashton-Miller, J.~O. DeLancey, J.~Luo, Convolutional neural
  network-based pelvic floor structure segmentation using magnetic resonance
  imaging in pelvic organ prolapse, Medical Physics.

\bibitem{rapacchi2019simultaneous}
S.~Rapacchi, T.~Troalen, Z.~Bentatou, M.~Quemeneur, M.~Guye, M.~Bernard,
  A.~Jacquier, F.~Kober, Simultaneous multi-slice cardiac cine with
  fourier-encoded self-calibration at 7 tesla, Magnetic Resonance in Medicine
  81~(4) (2019) 2576--2587.

\bibitem{makki2021new}
K.~Makki, A.~Bohi, A.~C. Ogier, M.-E. Bellemare, A new geodesic-based feature
  for characterization of 3d shapes: application to soft tissue organ temporal
  deformations, in: 2021 25th International Conference on Pattern Recognition,
  IEEE, 2021.

\end{thebibliography}
%%%%%%%%%%%%%%%%%%%%%%%%%%%%%%%%%%%%%%%%%%%%%%%%%%%%%%%%%%%%%%%%%%%%%%%%%

%%%%%%%%%%%%%%%%%%%%%%%%%%%%%%%%%%%%%%%%%%%%%%%%%%%%%%%%%%%%%%%%%%%%%%%%%
\newpage
\section*{Supporting materials caption}
Animations describing the volume deformations of 2 observed bladders. Global displacement is not reproduced, but colormap provided a visualization of the local surface changes. First column: reconstruction of the 3D dynamic volumes (blue) on the skeletons from the multi-slice acquisition segmentations (red). Second column: Projection on the static volume of the Jacobian maps associated with the registrations generating the reconstructed complete dynamic volumes. Third column: Single projection of the $\sigma_{J}$ map on the static volume. Volumes from each geometric configuration were placed in a common time frame but the volume rate per second is different and each configuration was imaged during independent acquisition. For each representation, the anterior (left) and posterior (right) views are provided.
%%%%%%%%%%%%%%%%%%%%%%%%%%%%%%%%%%%%%%%%%%%%%%%%%%%%%%%%%%%%%%%%%%%%%%%%%

%%%%%%%%%%%%%%%%%%%%%%%%%%%%%%%%%%%%%%%%%%%%%%%%%%%%%%%%%%%%%%%%%%%%%%%%%
\end{document}